\PassOptionsToPackage{font=small}{caption}
\documentclass[10pt,twocolumn,letterpaper]{article}

\usepackage{iccv}
\usepackage{times}
\usepackage{epsfig}
\usepackage{graphicx}
\usepackage{amsmath}
\usepackage{amssymb}
\usepackage{subfig}
\usepackage{xcolor}

\usepackage[pagebackref=true,breaklinks=true,letterpaper=true,colorlinks,bookmarks=false]{hyperref}

\makeatletter
\renewcommand{\paragraph}{%
  \@startsection{paragraph}{4}%
    {\z@}{1.25ex \@plus 1ex \@minus .2ex}{-1em}%
    {\normalfont\normalsize\bfseries}%
}
\makeatother

\iccvfinalcopy 

\newcommand\blfootnote[1]{%
  \begingroup
  \renewcommand\thefootnote{}\footnote{#1}%
  \addtocounter{footnote}{-1}%
  \endgroup
}

\begin{document}

\title{Lifting GIS Maps into Strong Geometric Context for Scene Understanding 
}
\author{Ra{\'u}l D{\'i}az, Minhaeng Lee, Jochen Schubert, Charless C. Fowlkes\\
Computer Science Department, University of California, Irvine\\
{\tt\small \{rdiazgar,minhaenl,j.schubert,fowlkes\}@uci.edu}
}

\maketitle
\begin{abstract}
Contextual information can have a substantial impact on the performance of
visual tasks such as semantic segmentation, object detection, and geometric
estimation. Data stored in Geographic Information Systems (GIS) offers a rich
source of contextual information that has been largely untapped by computer
vision. We propose to leverage such information for scene understanding
by combining GIS resources with large sets of unorganized photographs
using Structure from Motion (SfM) techniques. We present a pipeline to quickly
generate strong 3D geometric priors from 2D GIS data using SfM models aligned
with minimal user input. Given an image resectioned against this model, we
generate robust predictions of depth, surface normals, and semantic labels. 
Despite the lack of detail in the model, we show that the precision of the
predicted geometry is substantially more accurate than other single-image depth
estimation methods. We then demonstrate the utility of these contextual
constraints for re-scoring pedestrian detections, and use these GIS contextual
features alongside object detection score maps to improve a CRF-based semantic
segmentation framework, boosting accuracy over baseline models.

\blfootnote{This work was supported by NSF IIS-1253538 and a Balsells Fellowship to RD.}
\end{abstract}

\vspace{-.5cm}

\section{Introduction} \label{Introduction}
The problems of object detection and estimation of 3D
geometry have largely been pursued independently in computer vision. However, there seem to be
many good arguments for why these two sub-disciplines should join forces. Accurate
recognition and segmentation of objects in a scene should constrain matching of
features to hypothesized surfaces, aiding reconstruction. Similarly, geometric
information should provide useful features and contextual constraints for object detection and
recognition. The use of detailed stereo depth has already proven to be
incredibly effective in the world of object detection, particularly for RGB-D
sensors in indoor scenes \cite{gupta2014learning,lin2013holistic}. More general formulations have attempted to
jointly integrate geometric reconstruction from multiple images with scene and
object recognition \cite{Bao12,Cornelis08,Hoiem08}. On the other hand,
works such as \cite{Hoiem05,Gupta11,Murphy03,Sun12} have focused on the role of
context in a single image, operating under the assumption that the camera and
scene geometry are unknown and must largely be inferred based on analysis of
monocular cues. This problem is quite difficult in general although some
progress has been made \cite{Hoiem05popup,Saxena07,eigen2014depth}, particularly on indoor
scenes of buildings where the geometry is highly regular
\cite{Satkin13,Hedau09,Hedau10}.

In this paper, we argue that a huge number of photographs taken in 
outdoor urban areas are pictures of known scenes for which rich 
geometric scene data exists in the form of GIS maps and other geospatial data 
resources. Robust image matching techniques make it feasible to resection a 
novel image against large image datasets to produce 
estimates of camera pose on a world-wide scale \cite{Li12}. Once a test photo 
has been precisely localized, much of this contextual information can be easily 
backprojected into the image coordinates to provide much stronger priors for 
interpreting image contents. For the monocular scene-understanding purist, 
this may sound like ``cheating'', but from an applications perspective, such 
strong context is already widely available or actively being assembled and 
should prove hugely valuable for improving the accuracy of image understanding.

\begin{figure*}[ht]
\begin{center}
\includegraphics[width=0.8\paperwidth]{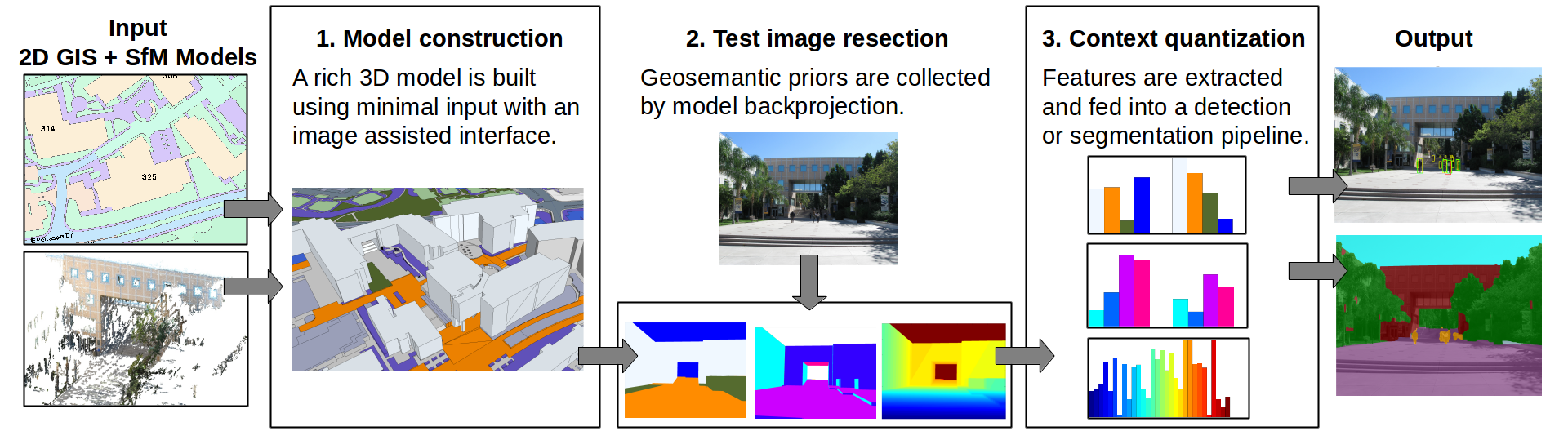}\\
\vspace{-.3cm}
   \caption{System overview: 2D GIS and SfM data are aligned to build a 3D model
with minimal effort using an image-assisted Sketchup plug-in. This 
fused geocontext provides a basis for efficiently transferring rich geometric and 
semantic information to a novel test image where it is used to improve performance of general 
scene understanding (depth, detection, and segmentation).}
    \label{fig:pipeline}
    \vspace{-0.1in}
\end{center}
\end{figure*}

To study the role of strong geometric context for image understanding, we have 
collected a new dataset consisting of over six thousand images covering a 
portion of a university campus for which relative camera pose and 3D 
coordinates of matching points has been recovered using structure from motion. 
We describe a method for aligning this geometric and photometric data with 2D 
GIS maps order to quickly build 3D polygonal models of buildings, sidewalks, 
streets and other static structures with minimal user input (Section \ref{Model}). 
This combined geosemantic context dataset serves as a reference against which 
novel test images can be resectioned, providing a rich variety of geometric and 
semantic cues for further image analysis. We develop a set of features for use 
in detection and segmentation (Sections \ref{Detection} and \ref{Segmentation}) 
and experimentally demonstrate that these contextual cues offer significantly 
improved performance with respect to strong baselines for scene depth 
estimation, pedestrian detection and semantic segmentation (Section 
\ref{Results}).


\section{Contribution and Related Work} \label{Related}

The contribution of our work is twofold. First, we demonstrate a pipeline that
performs precise resectioning of test images against GIS map data to generate
geosemantic context applicable to multiple scene understanding tasks. Secondly,
we show that simple cues derived from the GIS model can provide significantly
improved performance over baselines for depth estimation, pedestrian detection
and semantic segmentation. We briefly discuss the differences between our
results and closely related work. Figure \ref{fig:pipeline} shows an overview of
the system pipeline.

\paragraph{GIS for image understanding} 
The role of GIS map data in automatically interpreting
images of outdoor scenes appears to have received relatively little attention
in computer vision. While detailed GIS map data is used extensively in
analysis of aerial images \cite{Uchiyama09,Wang09,rumpler2012rapid}, only a
handful of papers have exploited this resource to study scene understanding
from a ground-level perspective. Recently, a few groups have looked at using
GIS data and multi-view geometry for improving object recognition.
\cite{Matzen13} introduced a geographic context re-scoring scheme for car
detection based on street maps. Ardeshir et al. used GIS data as a prior for
static object detection and camera localization \cite{Ardeshir14} and for
performing segmentation of street scenes \cite{ardeshir2015geo}. Compared to
these works, our pipeline utilizes full 6D camera pose estimation and a richer  
scene model derived from the GIS map that supports not only improved detection  
rescoring but also depth and semantic label priors.

Perhaps most closely related work to our approach is \cite{Wang_2015_CVPR}
which uses a CRF to simultaneously estimating depth and scene labels using
a strong 3D model. However, \cite{Wang_2015_CVPR} assumes that
both camera localization and a CAD model of a scene with relevant categories
(e.g. trees) are provided as inputs. In contrast, we address the construction
of a 3D scene model by combining image and map data as well as test-time camera
localization using model alignment and resectioning techniques. Interestingly,
we also show that lifted GIS data can improve segmentation accuracy even for
semantic labels that are not present in the GIS map model (e.g. trees,
retaining walls).


\paragraph{Contextual detection rescoring}
Rescoring object detector outputs based on contextual and geometric scene
constraints has been suggested in a wide variety of settings. For example,
Hoiem et al. \cite{Hoiem06} used monocular estimation of a ground-plane to
rescore car and pedestrian detections, improving the performance of a contemporary
baseline detector \cite{dalal2005histograms}. Similarily, \cite{Ardeshir14}
showed improvement on hard to detect objects such as fire hydrants.
However, the benefit of scene geometry constraints appears to be substantially
less when a more robust baseline detector such as DPM \cite{Felzenszwalb10} is
used. For example, \cite{Diaz13} reported that monocular ground-plane
constraints failed to improve the performance of a DPM pedestrian detector.
Similarly, \cite{Matzen13} reported that geometric rescoring based on road maps
did not improve performance of a DPM car detector (although rescoring did
improve detection for a weaker detector that had been trained to predict
viewpoints). In contrast to this previous work, we use a richer rescoring
model that allows for non-flat supporting surfaces and integrates additional
geosemantic cues, resulting in a significant boost (5\% AP) in pedestrian
detection even when rescoring a strong baseline model. 

%
%

\section{Lifting GIS Maps} \label{Model}

In this section, we describe the construction of a dataset for exploring 
the use of strong GIS-derived geometric context. We focus on novel aspects of
this pipeline which include: aligning GIS and SfM data, a user-friendly toolbox
to lift 2D geosemantic information provided by a map into 3D geometric context
models, and model-assisted camera resectioning to allow quick and
accurate localization of a test image with respect to the context model.

\paragraph{Image database acquisition}
We collected a database of 6402 images covering a large area (the engineering
quad) of a university campus. Images were collected in a systematic manner
using a pair of point and shoot cameras attached to a monopod. We
chose locations so as to provide approximately uniform coverage of the area of interest.
Images were generally collected during break periods when there were relatively
few people present (although some images still contain pedestrians and other
non-rigid objects).

Running off-the-shelf incremental structure from motion (e.g.,
\cite{Agarwal09,Snavely07}) on the entire dataset produces a 3D structure that
is qualitatively satisfying but often contains metric inaccuracies. In
particular, there can be a significant drift over the whole extent of the 
recovered model that makes it impossible to globally align the model with GIS
map data. These reconstruction results were highly dependent on the
quality of the initial matches between individual camera pairs. However, we found that the
excellent incremental SfM pipeline implementation in OpenMVG~\cite{openMVG},
which uses the adaptive thresholding method of \cite{moulon2013adaptive},
yielded superior results in terms of the accuracy and number of recovered
cameras and points. Our final model included 4929 succesfully bundled images.

\paragraph{Global GIS-structure alignment}

We obtained a 2D GIS map of the campus maintained by the university's
building management office. The map was originally constructed from aerial imagery
and indicates polygonal regions corresponding to essential campus 
infrastructure tagged with semantic labels including building footprints,
roadways, firelanes, lawns, etc. 

We would like to align our SfM reconstruction with this model. One approach is to
leverage existing sets of geo-referenced images. For example, \cite{Matzen13}
used aerial LiDAR and Google Street View images with known geo-coordinates in
order to provide an absolute coordinate reference. While this is practical
when the 3D GIS data has already been generated, we start from a flat 2D model
depicting an environment for which no precisely georegistered images were
readily available.

To quickly produce an initial rough alignment, we project the point cloud
recovered using SfM onto a ground-plane and have a user select 3 or more points
(typically building corners as they are more visible). We run Procrustes
alignment to match the user selected points to the real coordinates in our 2D
GIS data. This global 2D alignment is sufficient to register the SfM model in
geographic coordinates for the intial construction of a 3D model. Once the 3D GIS
model is defined (see below), we used an iterative closest point approach to 
automatically refine the alignment of the GIS model and SfM point cloud.

\paragraph{Image-assisted 3D model construction}

\begin{figure}[t]
\begin{center}
\includegraphics[width=0.8\linewidth]{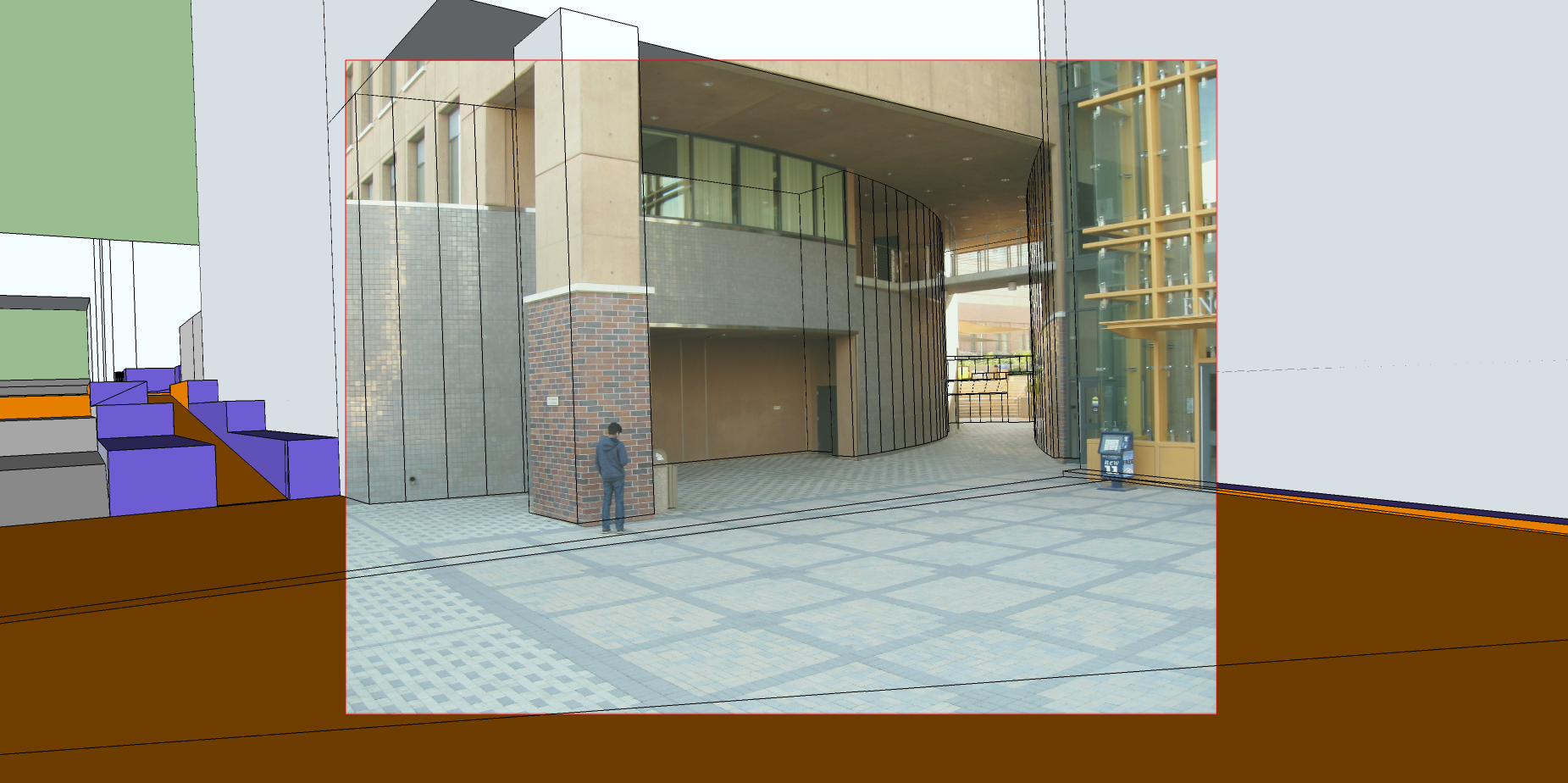}
\end{center}
\vspace{-.5cm}
   \caption{We developed a custom Sketchup plug-in that imports camera parameters 
   computed during bundle adjustment. The user easily models the 3D geometry by extruding,
   tilting, and carving the 2D map data until it naturally aligns with the image.
}
\label{fig:sketchup}
\vspace{-0.2in}
\end{figure}

We developed a custom plug-in for the 3D modeling tool Sketchup \cite{sketchup}
to allow efficient user-assisted ``lifting'' of 2D GIS map data into a full 3D
model. We imported the GIS 2D polygons with their corresponding semantic labels into
the workspace. The user is then presented with a choice of images to load from the
globally aligned SfM model. When an image is selected, the 3D model view is adjusted
to match the recovered camera extrinsic and intrinsic parameters and the image is
transparently overlayed, providing an intuitive visualization as shown in Figure \ref{fig:sketchup}.
The user can then extrude flat 2D polygons (e.g., building footprints) to the
appropriate height so that the backprojected view into the selected camera
matches well with the overlayed image.

A full 3D mesh model can be easily constructed starting from the aligned 2D map by 
extruding buildings up, carving stairs down, tilting non-fully
horizontal surfaces, etc. using standard Sketchup modeling tools and guided by
the image overlay. Additional geometry can be easily created, as well as
adding additional semantic labels or corrections to the original data. With the
assistance of these aligned camera views, constructing a fairly detailed model
in Sketchup covering 10 buildings took approximately 1-2 hours. This can be
considered an offline task since the modeling effort is performed only once, as
buildings are largely static over time.

\paragraph{GIS-assisted test image resectioning}
To estimate camera pose at test time, we resection each test image against the
SfM image dataset based on 2D-3D matching correspondences using a
RANSAC-based 3-point absolute pose procedure \cite{kneip2011novel}. We
developed several techniques for improving the accuracy of this resectioning.
As the bundled image dataset grows to cover larger areas, resectioning accuracy
generally falls since the best match for a given feature descriptor in the test
image is increasingly likely to be a false positive. To leverage knowledge of
spatial locality and scale to large datasets, we partitioned the bundled
cameras into $k=10$ clusters using k-means over database camera positions.
Points in the bundled model were included in any camera cluster in which they
were visible yielding $10$ non-disjoint clusters of points.

Each test image was resectioned independently against each spatial cluster and
the best camera pose estimate among the clusters was selected using geometric
heuristics based on the GIS model. We measured the distance to ground (height) and
camera orientation with respect to the gravity direction in the GIS model. All
camera poses below the ground plane, higher than 4 meters, or tilted more than
30 degrees were discarded. If a camera pose was geometrically reasonable in
more than one cluster, we selected the estimate with the highest number of
matched inliers. {\em Incorporating these heuristic constraints increased the
proportion of correctly resection test images substantially (from 49\% to
59\%).} This improvement would not possible without the GIS-derived 3D model
since the ground elevation varies significantly and cannot be captured by a
simple global threshold on the camera coordinates.

Resectioning of a test image produces immediate predictions of scene depth,
surface orientation, and semantic labels at every image pixel. It is
interesting to compare these to previous works that attempt to estimate such
geometric or semantic information from a single image. By simply resectioning
the picture against our 3D model we are immediately able to make surprisingly
accurate predictions without running a classifier on a single image patch! In
the remainder of the paper, we discuss how to upgrade these ``blind''
predictions by incorporating backprojected model information into standard
detection and segmentation frameworks.

\section{Strong Geometric Context for Detection} \label{Detection}

Estimating camera pose of a test image with respect to a 3D GIS model aids
in reasoning about the geometric validity of hypothesized object detections
(e.g. pruning false positives in unlikely locations or boosting low-scoring
candidates in likely regions). We describe a collection of features that
capture these constraints and use them to improve detection performance. We
incorporate these features into a pedestrian detector by training an SVM with
geometric context features (GC-SVM) that learns to better discriminate object
hypotheses based on the geosemantic context of a candidate detection. 

\paragraph{3D geometric context} 
Let a candidate 2D bounding box in a test image $I$ have an associated
height in pixels $h^{im}$. We use a deformable part model that consists of a
mixture over three different template filters: full-body, half upper body, and
head. We set the image detection height $h^{im}$ based on which mixture fires
as $1$, $2$ or $3$ times the bounding box height respectively.

If we assume that the object is resting on a horizontal surface and the base of
the object is visible, then we can estimate the 3D location the object is
occupying. We find the depth at intersection $z_i$ of the object with respect to the camera by
shooting a ray from the camera center through the bounding box's ``feet'' and
intersecting it with the 3D model. Importantly, unlike many previous works,
the ground is not necessarily a plane (e.g., our model includes ground at
different elevations as well as stairs and ramps). Given camera focal length $f$, 
we can estimate the height in world coordinates by the following expression:
\begin{equation} 
\label{eq:Wh}
h_i = \frac{z_i}{f} h^{im}
\end{equation}

Unfortunately, the object's ``feet'' might not be visible at all times (e.g. a low 
wall is blocking them). Let $h_{\mu}$ be the ``physical height'' of an average
bounding box (i.e., the height of an average human). We collect all possible intersections with
the model (a blocking wall and the ground plane hehind it) and choose the $h_i$ that
minimizes $(h_i - h_{\mu})^{2}$.

An alternative to hypothesize an object's height is by its inverse relation with depth.
We can estimate an object's depth $z_o$ based on the expected average human height $h_{\mu}$. 
A bounding box of size $h^{im}$ in the image has an expected distance
\begin{equation} 
z_o = h_{\mu} \frac{f}{h^{im}}
\end{equation}
from the camera.  Given $z_o$, we produce a second height estimate $h_o$ by
tracing a ray through the center top of the detection to a depth $z_o$ and then
measuring the distance from this estimated head position to the ground plane.

For each height estimator $h_i,h_o$ we also extract a corresponding semantic
label associated with the GIS-model polygon where the feet intersect and record
binary variables $w_i,w_o$ indicating whether the polygon has a ``walkable''
semantic label and $n_i,n_o$ indicating wheter it has a horizontal surface
normal.  Our feature vectors for each estimate are given by:
\begin{align}
F_i &= [v_i(h_i-h_\mu)^2, w_i, n_i, (1-v_i)]\\
F_o &= [v_o(h_o-h_\mu)^2, w_o, n_o, (1-v_o)]
\end{align}
where $h_\mu = 1.7m$ is the average human height and $v_i,v_o$ are binary variables
indicating whether the the corresponding height cound be measured. For
example, if the ray to the foot to compute $z_i$ does not intersect the model or if the depth
estimate $z_o$ is behind the model surface, we mark the estimate as invalid and
zero out the feature vector.

\paragraph{2D geosemantic context}
In addition to the height and foot location, we extract geosemantic context by
backprojecting model semantic labels and surface normals into the image plane
and look at the distribution inside of the object bounding box. For each bounding
box $b$, we hypothesize a full-body bounding box given the detection's mixture
as previously mentioned. We then split such box vertically into 3 parts (top,
center, bottom) and collect normalized histograms $H_{b}$ of the distribution
of semantic labels and surface normals in each subregion. We account for $5$
GIS labels (building, plants, pavement, sky, unknown) and $4$ discretized
surface normal directions (ground, ceiling, wall, none).
\begin{equation}
H_{b} = [H_{top},H_{center},H_{bottom}]
\end{equation}
To allow the learned SVM weights to depend on the original detector mixture
component $m$, we contruct an expanded feature vector from these histograms:
\begin{equation}
F_{b} = [\delta(m = 1)H_{b},\delta(m = 2)H_{b},\delta(m = 3)H_{b}]
\end{equation}
where $m \in 1,2,3$ indicates a full-body, half-upper, or head mixture
respectively. For example, this allows us to model that upper-body detections
are correlated with the presence of some vertical, non-walkable surface such as
a wall occluding the lower body.

We train an SVM to rescore each detection using a final feature vector that
includes the detector score $s$ along with the concatenated context features
\[
F = [s, F_i, F_o, F_b]
\]

\section{Strong Geometric Context for Segmentation} \label{Segmentation}

The geometric and semantic information contained in the GIS data and lifted
into the 3D GIS model can aid in reasoning about the geometric validity of
class hypotheses (e.g., horizontal surfaces visible from the ground are
not typically buildings). We describe methods for using such constraints to
improve semantic segmentation performance. We follow a standard CRF-based
labeling approach from Gould et al. \cite{gould2009decomposing} which uses an 
augmented set of image features from
\cite{shotton2006textonboost}. We explore simple ways to enhance this set of
features using GIS data and study its influence on semantic labeling
accuracy.

\paragraph{GIS label prior distributions} The GIS-derived model provides an
immediate estimation of pixel labels based on the 4 semantic labels in the original
GIS map (building, plants, pavement, and sky). If a camera pose is known, we can backproject the model into the
image plane and transfer the polygon label in the GIS model to the projected
pixel. However, camera pose estimation is not perfect and might contain
minimal deviations from its ground truth pose. In order to account for slight
camera pose inaccuracies, we define a 16-dimensional feature descriptor to softly handle
these cases. Given an image $I$, a pixel $x \in I$, and a backprojected GIS
semantic label $g(x)$, we define the feature $h_{k}^{r}(x)$ 

\begin{equation} \label{eq:hg_eq}
h_{r,k}^{s}(x) = \frac{1}{N} \sum_{y:\|y-x\|<r}{[g(y) = k]}
\end{equation}
as the normalized count of class $k$ pixels in a circular disc of radius $r$
around $x$, where $N$ is the number of pixels in the disc. In our experiments,
we define $r$ so that the angular error of the camera pose is 0, 1, 3, and 5
degrees.

\paragraph{GIS surface normal distributions} In a similar manner, a surface
normal can be quickly estimated for any pixel by backprojecting the 3D model
into the camera plane. Surface normals can be discriminative of certain classes
like pavement, roads, buildings, etc. Following the same structure as in
equation \ref{eq:hg_eq}, we define the 12-dimensional feature $h_{n}^{r}(x)$ as
\begin{equation} \label{eq:hg_eq2}
h_{r,k}^{n}(x) = \frac{1}{N} \sum_{y:\|y-x\|<r}{[n(y) \in N_k]}
\end{equation}
where $N_k$ is one of 3 possible surface orientation bins: horizontal (ground),
horizontal (ceiling), vertical (wall).

\begin{figure}[t]
\captionsetup[subfigure]{labelformat=empty}
\centering
\def\arraystretch{0.3}
\subfloat[Ground Truth]{
\hspace{-0.5cm}
\begin{tabular}{c}
        \includegraphics[width=0.25\columnwidth]{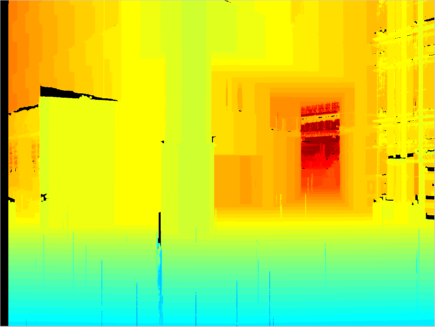} \\
        \includegraphics[width=0.25\columnwidth]{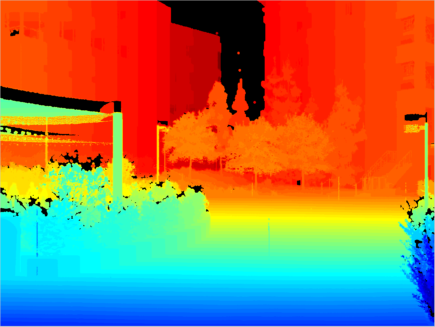} \\
        \includegraphics[width=0.25\columnwidth]{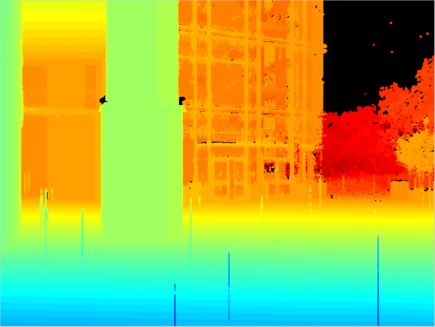} \\
\end{tabular} 
}
\subfloat[Make3D \cite{Saxena07}]{
\hspace{-0.6cm}
\begin{tabular}{c}
        \includegraphics[width=0.25\columnwidth]{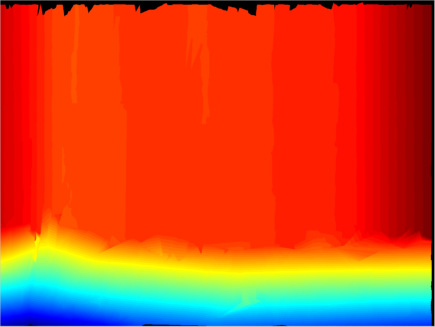} \\
        \includegraphics[width=0.25\columnwidth]{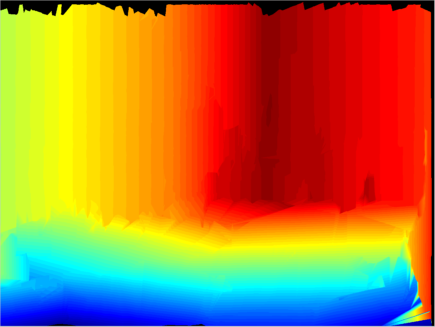} \\
        \includegraphics[width=0.25\columnwidth]{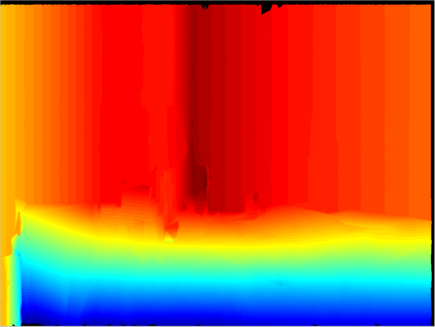} \\
\end{tabular} 
}
\subfloat[DNN \cite{eigen2014depth}]{
\hspace{-0.6cm}
\begin{tabular}{c}
        \includegraphics[width=0.25\columnwidth]{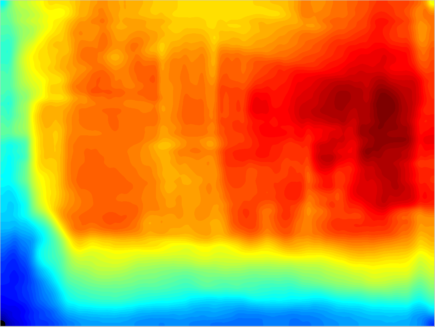} \\
        \includegraphics[width=0.25\columnwidth]{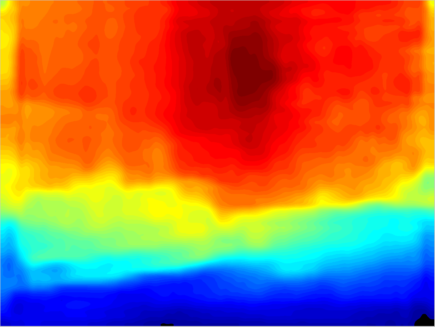} \\
        \includegraphics[width=0.25\columnwidth]{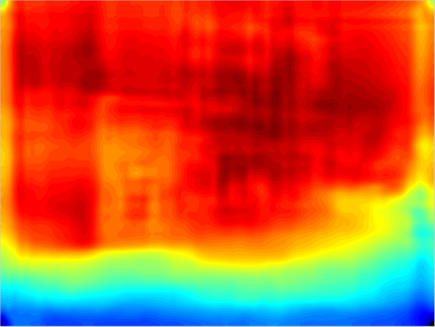} \\
\end{tabular} 
}
\subfloat[GIS]{
\hspace{-0.6cm}
\begin{tabular}{c}
	\includegraphics[width=0.25\columnwidth]{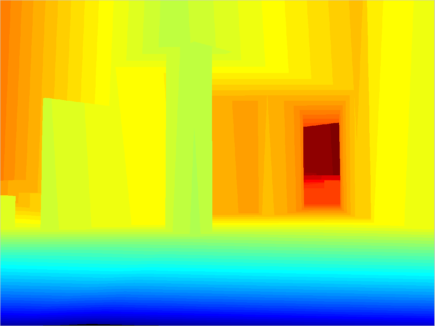} \\
	\includegraphics[width=0.25\columnwidth]{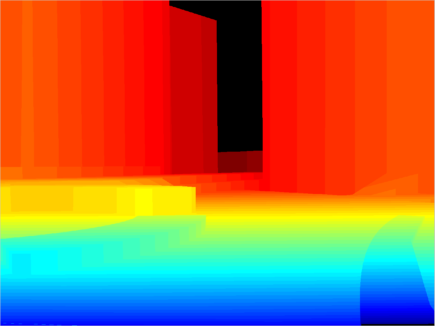} \\
	\includegraphics[width=0.25\columnwidth]{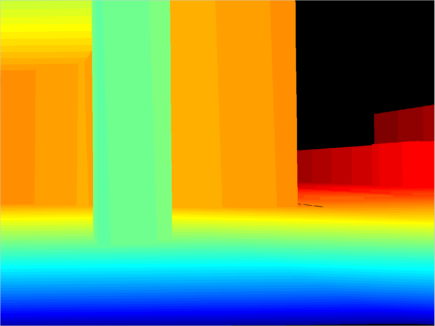} \\
\end{tabular} 
}
   \caption{Qualitative depth comparison. Our GIS backprojected depth map is
   shown in the last column. While it lacks many details such as foliage 
   and pedestrians which are not included in our coarse GIS-based 3D model,
   simply backprojecting depth provides substantially more accurate estimation
   than existing monocular approaches.}
\label{fig:det_results}
\vspace{-0.2in}
\end{figure}

\paragraph{GIS Depth features} Depth can also be efficiently estimated when
from a 3D model when a camera pose is known. Following other methods like
\cite{yang2014action,rogez20143d}, we extract HOG features
\cite{dalal2005histograms} to encode depth variations. We did find a substantial gain 
in certain categories when adding these features into the model (e.g. wall).

\paragraph{DPM as a context feature} Inspired by other works that try to create
segmentation-aware detectors \cite{fidler2013bottom,ladicky2010and,martinovic2012three,riemenschneider2012hough}, we also incorporate the
outputs of category-specific object detectors in our segmentation model. To do
so, we collect the scores of a DPM detector for an object category $c$ and
generate a DPM feature map $h_{c}$ by assigning to every pixel the maximum
score of any of the candidate detection boxes intersecting the given pixel.
Let $\Omega_{c}$ be the set of candidate detections and $b_i, s_i$ the bounding
box and score for the $i$th detection, then
\begin{equation} \label{eq:obj_eq}
h_{c}^{o}(x) = \max_{i \in \Omega_{c}}(s_i \cdot [x \in b_i] )
\end{equation}

\section{Experimental results} \label{Results}

In our experiments, we started with a test set comprising 570 pictures taken in
the area covered by the 3D model. These images were collected over different
days and several months after the initial model dataset was acquired. Of these
images, 334 images (59\%) were successfully resectioned using the cluster-based
approach from Section \ref{Model}. This success rate compares favorably with
typical success rates
for incremental bundle adjustment (e.g., \cite{Matzen13} register 37\% of their
input images) even though our criteria for correctness is more stringent
(we manually scored test images rather than relying on number of matching feature points).
 To evaluate performance of detection, we annotated resectioned
images with ground-truth bounding boxes for 1484 pedestrians 
using the standard PASCAL VOC labeling practice including tags for truncated and
partially occluded cases. We used 167 images for training our geometric context rescoring 
framework and left the remaining 167 for testing. We also manually segmented 305 of these images 
using the segmentation tool provided by \cite{MYP:BMVC:2013} and labeled each segment
with one of 9 different semantic categories. We split the segmentation data
into 150 images for training and 155 for testing.

\begin{figure}[t]
\begin{center}
{\scriptsize
\begin{tabular}{|c||c|c|c|} 
\hline
Threshold            & Make3D  & DNN     & GIS     \\ \hline
$\delta < 1.25$      & 11.03\% & 28.21\% & 67.04\% \\ \hline
$\delta < 1.25^{2}$  & 30.71\% & 45.09\% & 82.37\% \\ \hline
$\delta < 1.25^{3}$  & 49.47\% & 56.04\% & 88.25\% \\ \hline
\end{tabular}
}
\end{center}

\vspace{-.2cm}
   \caption{Quantitative comparison of depth estimation methods: Make3D
   \cite{Saxena07}, Deep Neural Network \cite{eigen2014depth}, and GIS
   backprojection. We list the proportion of depths within a specified maximum 
   allowed relative error $\delta = \max({\frac{d_{gt}}{d_{est}},\frac{d_{est}}{d_{gt}}})$, where $d_{gt}$ is the
   ground truth depth and $d_{est}$ is the estimated depth at some point. 
   The DNN model predictions were re-scaled to match our ground truth 
   data since the model provided is adapted only for indoor scenes.
}
\label{fig:dt-table}
\vspace{-0.2in}
\end{figure}

\subsection{Monocular Depth Estimation}

To verify that the coarse-scale 3D GIS model provides useful geometric
information, despite the lack of many detailed scene elements such as trees, we
evaluated resectioning and backprojection as an approach for monocular depth
estimation. While our approach is not truly monocular since it
relies on a database of images to resection the camera, the test time data
provided is a single image and constitutes a realistic scenario for monocular
depth estimation in well photographed environments.

To establish a gold-standard estimate of scene depth, we scanned 14 locations
of the area covered by our dataset using a Trimble GX3D terrestrial laser
scanner. We took the scans at in a range of resolution between 5 and 12 cm in
order to keep the total scanning time manageable, resulting in roughly a half a
million 3D points per scan. We mounted a camera on top of the laser scanner
and used the camera focal length to project the laser-based 3D point cloud onto
the camera image plane, interpolating depth values to obtain a per-pixel depth
estimate. We then resectioned the test image and synthesized the depth map
predicted by our 3D GIS model.

Figure~\ref{fig:dt-table} shows quantitative results of our GIS backprojection
depth estimation against other single-image depth approaches for the 14 scan
dataset. We used the provided pre-trained models included in
\cite{Saxena07,eigen2014depth} as baselines for comparison. Since the pretrained DNN
model \cite{eigen2014depth} was only available for indoor scenes (trained from
Kinect sensor data), we estimated a scaling and offset factors for the output
that minimized the relative error over the 14 images. While the optimally
scaled pretrained DNN model is probably suboptimal, it is not possible to
retrain the model without collecting substantially larger amounts of training
data using specialized hardware (a laser scanner).  In contrast, our proposed
approach of simply backprojecting GIS data greatly outperforms
image-predictions using only camera pose and a map, with no training data
required!

\begin{figure}[t]
\begin{center}
 \includegraphics[width=0.7\linewidth]{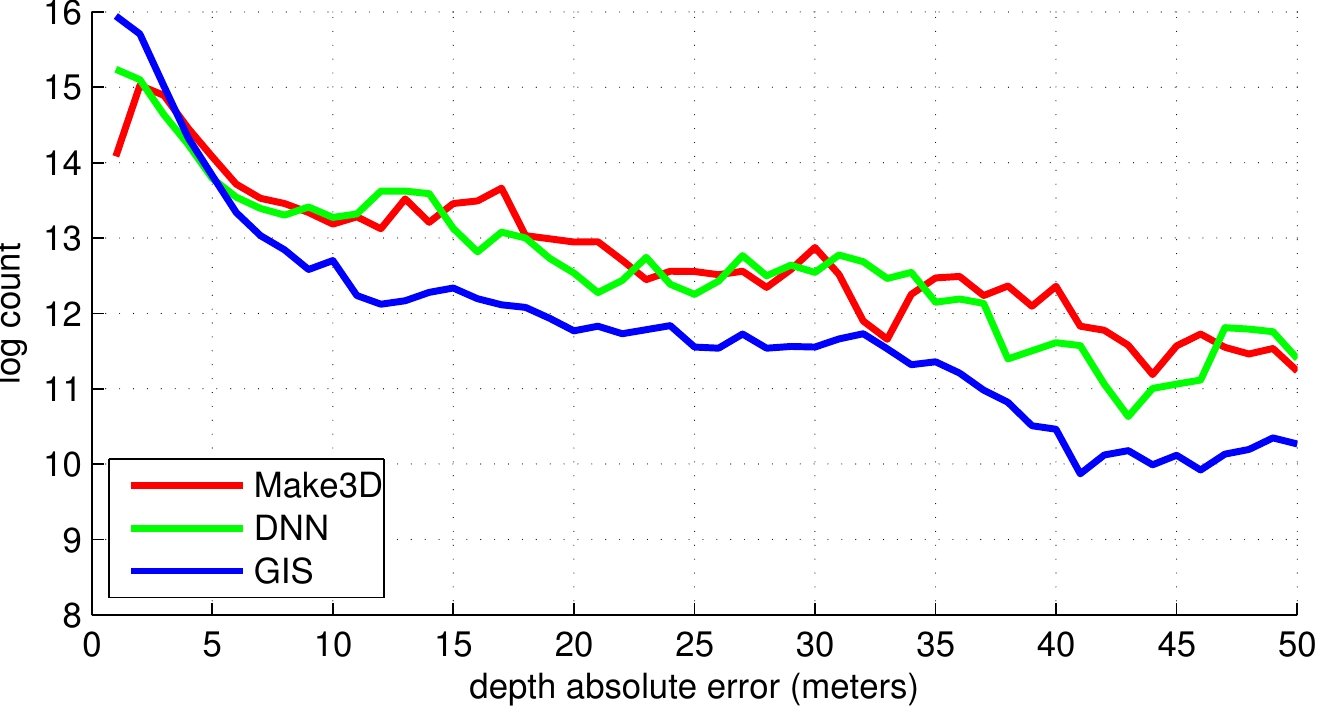} \\
 \includegraphics[width=0.7\linewidth]{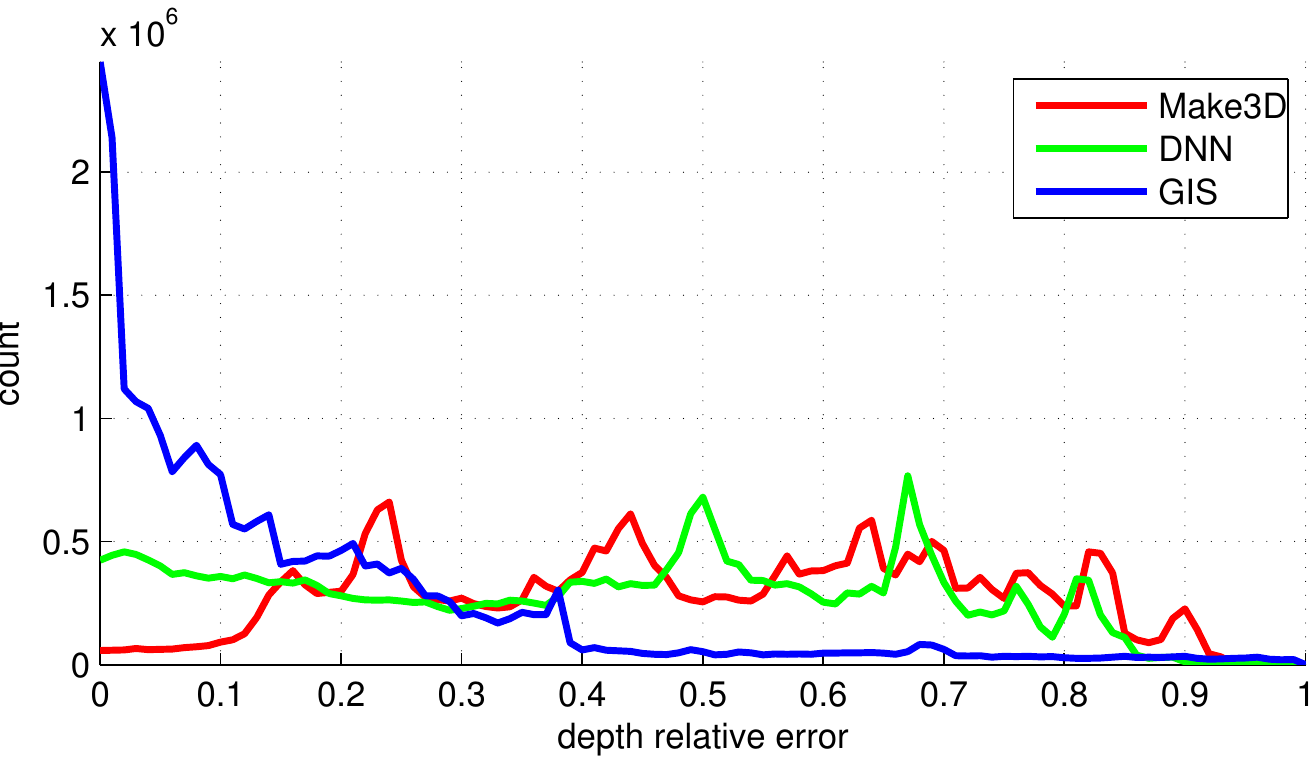} \\
   \caption{Accuracy of predicted depth estimates compared to gold-standard provided
   by a laser-scanner for 14 images. Top histogram shows the distribution (log
   counts) of absolute depth errors between Make3D, DNN and depth computed from
   resectioning and backprojecting GIS derived 3D model. The bottom plot shows
   the distribution of relative error $|d_{est} - d_{gt}|/d_{gt}$.
   \label{fig:lidarhist}
}
\end{center}
\vspace{-0.2in}
\end{figure}

\subsection{Object Detection}

We evaluated our geometric and semantic object rescoring scheme applied to the
widely used deformable part model detector (DPM) \cite{Felzenszwalb10} implemented in
\cite{voc-release5}. We used a standard 
non-maxima suppression ratio of 0.5 in the intersection over 
union overlap.

Of the 167 training image split, we collected the set of geometric context 
features $F$ described in Section \ref{Detection} along with the appropriate 
label indicating whether the bounding box was a true or false positive. 
We trained a linear SVM classifier using the implementation of \cite{libsvm}.
In our experiments, we set the regularization $C=4$ using 5-fold
cross-validation.  To accommodate class imbalance and maximize 
average precision, we used
cross-validation to set the relative penalty for misclassified positives to be
$4x$ larger than for negatives.

We benchmarked the trained Geometric Context SVM classifier using the same
features $F$ collected for the 167 image test split. While standard
DPM detector score provided a baseline average precision (AP) of \textbf{0.457},
our GC-SVM model achieved an AP of 
\textbf{0.507}. 
It is important to note that previous attempts to incorporate GIS-based
geometric rescoring into a DPM classifier provided little to no improvement.
In \cite{Matzen13}, 3D context between cars and streets allowed for improved
geometric reasoning about car orientation but only small gains in detection
performance. In fact their final VP-LSVM and VP-WL-SSVM models had lower
average precision than a baseline DPM model.

\begin{figure}[t]
\begin{center}
 \includegraphics[width=0.65\linewidth]{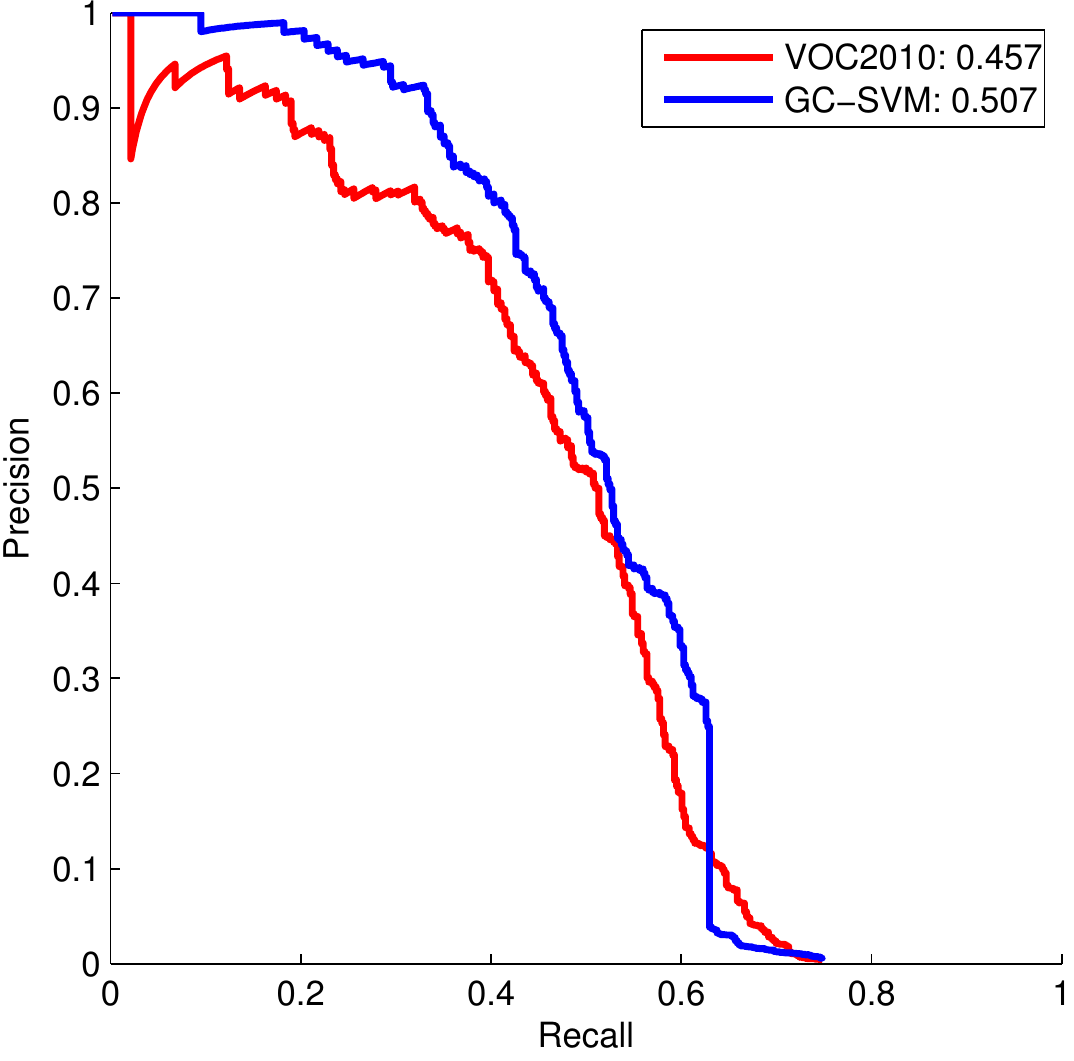}
   \caption{Geometric context aids in recognizing discriminative 3D and 2D features
that improve the average precision in pedestrian detection. Our GC-SVM obtained a
$5\%$ boost in AP with respect to the standard DPM model.
   \label{fig:prcurves}
}
\end{center}
\vspace{-0.2in}
\end{figure} 

\begin{figure}[h]
\captionsetup[subfigure]{labelformat=empty}
\centering
\def\arraystretch{0.3}
\setlength{\tabcolsep}{0.3mm}
\subfloat[DPM]{%
\hspace{-0.6cm}
\begin{tabular}{c}
	\includegraphics[width=0.48\columnwidth]{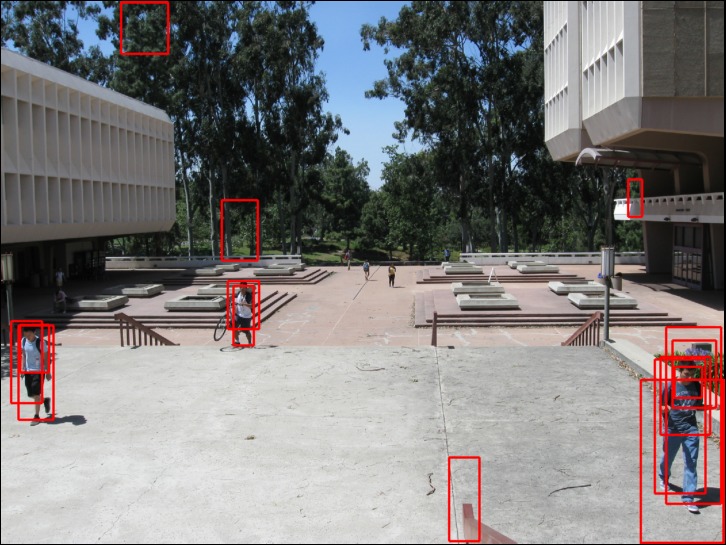} \\
	\includegraphics[width=0.48\columnwidth]{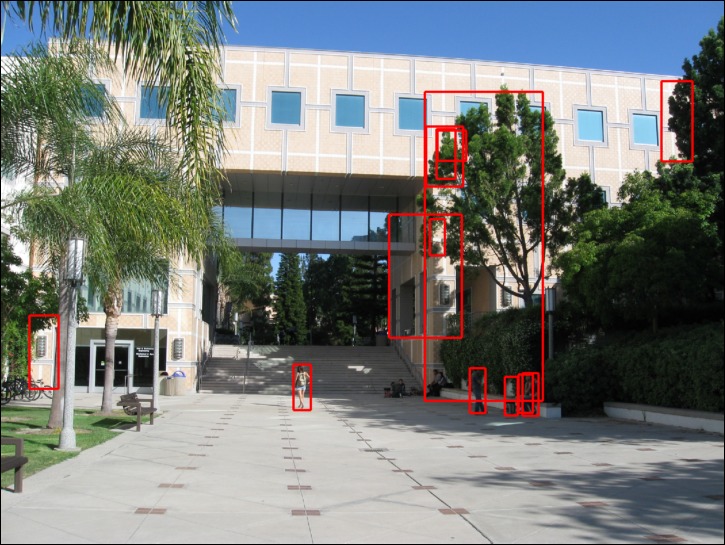} \\
	\includegraphics[width=0.48\columnwidth]{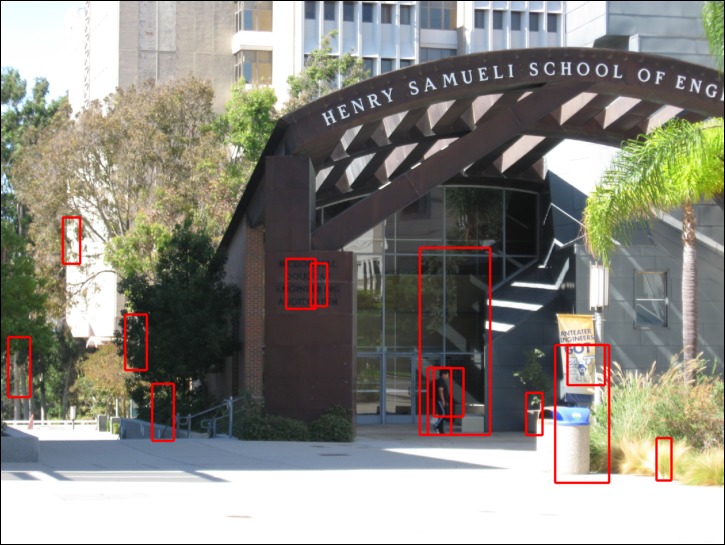} \\
\end{tabular} \hspace{-0.5cm}
}
\subfloat[GC-SVM]{%
\hspace{0.3cm}
\begin{tabular}{c}
	\includegraphics[width=0.48\columnwidth]{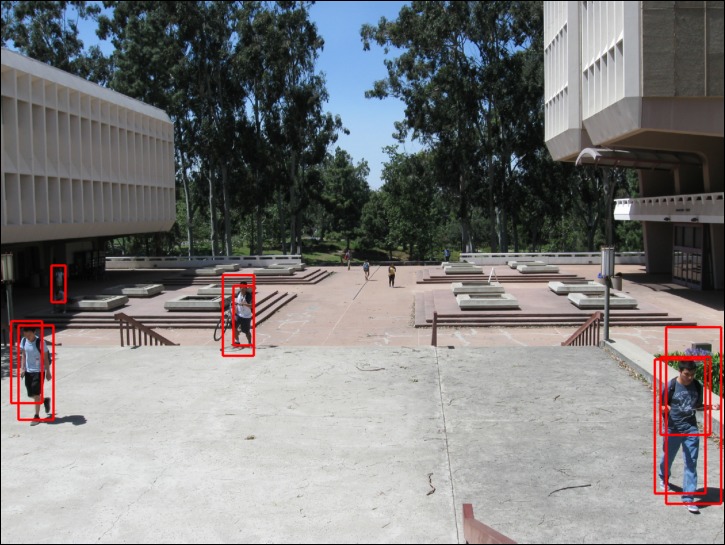} \\
	\includegraphics[width=0.48\columnwidth]{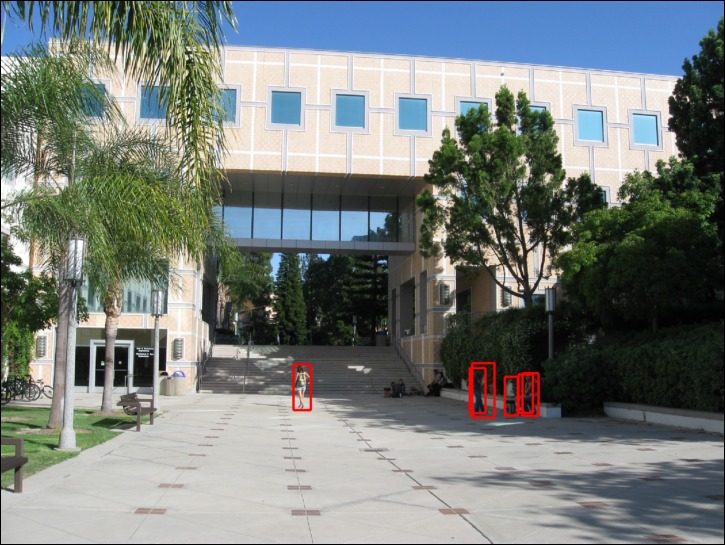} \\
	\includegraphics[width=0.48\columnwidth]{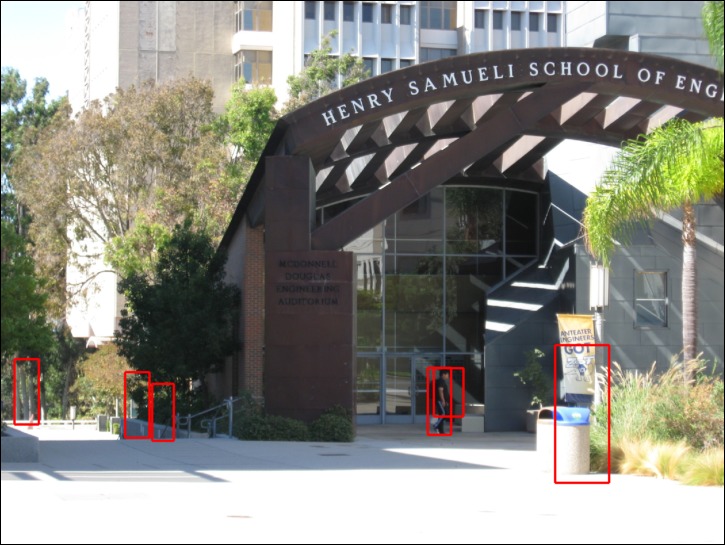} \\
\end{tabular} \hspace{-0.5cm}
}

\caption{Detection results at 0.6 recall. Geosemantic context successfully removes high DPM score
	false positives at unlikely places without adding too many low DPM score ones at coherent
	regions. This successful trade-off benefits the performance at almost all levels of
	recall.
}

\label{fig:det_results}
\vspace{-0.2in}
\end{figure}

\begin{figure*}[ht]
\begin{center}
{\scriptsize
\begin{tabular}{|l||c||c|c|c|c||c|c|c||c|c|}
\hline
Model & Overall & building & plants & pavement & sky & ped. & ped. sit & bicycle & bench & wall   \\ \hline
\hline
Image CRF (SUN)       & 0.309 & 0.767 & 0.581 & 0.863 & 0.872 & 0.007 & 0.000 & 0.000 & 0.000 & 0.000 \\ \hline
+DPM      & \textbf{0.323} & \textbf{0.774} & \textbf{0.586} & \textbf{0.864} & 0.873 & \textbf{0.129} & 0.000 & \textbf{0.001} & 0.000 & 0.000 \\ \hline
\hline
GIS Label Backprojection & 0.242 & 0.688 & 0.099 & 0.810 & 0.581 & 0.000 & 0.000 & 0.000 & 0.000 & 0.000 \\ \hline
GIS CRF         & \textbf{0.290} & \textbf{0.730} & \textbf{0.316} & \textbf{0.847} & \textbf{0.705} & 0.000 & 0.000 & 0.000 & 0.000 & \textbf{0.014} \\ \hline
\hline
Image CRF (ENGQ)        & 0.561 & 0.917 & 0.886 & 0.925 & 0.949 & 0.400 & 0.010 & 0.370 & 0.241 & 0.348 \\ \hline
+GIS        & 0.584 & 0.937 & 0.894 & 0.936 & 0.963 & 0.394 & 0.060 & 0.385 & 0.208 & \textbf{0.481} \\ \hline
\quad Depth   & 0.569 & 0.935 & \textbf{0.895} & 0.937 & 0.957 & 0.390 & 0.011 & 0.358 & 0.179 & 0.455 \\ \hline
\quad Labels    & 0.575 & \textbf{0.938} & 0.892 & 0.933 & \textbf{0.966} & 0.374 & 0.064 & 0.366 & 0.221 & 0.419 \\ \hline
\quad Normals & 0.568 & 0.935 & 0.893 & 0.933 & 0.961 & 0.389 & 0.007 & 0.385 & 0.192 & 0.418 \\ \hline
+DPM        & 0.590 & 0.920 & 0.892 & 0.929 & 0.947 & \textbf{0.583} & 0.013 & 0.482 & 0.184 & 0.360 \\ \hline
+DPM+GIS    & \textbf{0.627} & 0.936 & 0.894 & \textbf{0.938} & 0.961 & 0.568 & \textbf{0.108} & \textbf{0.520} & \textbf{0.245} & 0.472 \\ \hline

\end{tabular}
}
\end{center}

\vspace{-.2cm}
   \caption{Quantitative segmentation results for models trained with generic
   (SUN) and scene specific (ENGQ) data. Accuracy is measured using 
   PASCAL intersection-over-union (IOU) protocol. Adding geosemantic
   (+GIS) and detection (+DPM) features outperformed the baseline models.
   Combining both methods gave the best overall results in the scene specific
   model, although some classes did not achieve their best accuracy individually.
   }
\label{fig:seg-table}
\vspace{-0.2in}
\end{figure*}

\subsection{Semantic Segmentation}

We built two baseline segmentation models using the CRF-based multi-class 
segmentation code provided by \cite{gould2012darwin}. We trained
one model using our training labeled set of engineering quad
images (CRF ENGQ) as a scene-specific model. We also trained a generic model using images
collected from the online SUN dataset \cite{xiao2010sun} by querying for
multiple categories in the SUN label set that are semantically equivalent to
the 9 categories labeled in our engineering quad dataset (CRF SUN). Finally, we added our 
GIS and DPM features to the ENGQ model and evaluated their effects on segmentation performance.

\paragraph{Detectors Improve Segmentation}

We collected DPM score features as described previously for two objects of
interest: pedestrian and bicycle. Figure \ref{fig:seg-table} shows the
influence of adding these features into the model (+DPM rows). We raised 
pedestrian segmentation accuracy from $0.400$ to $0.583$ and bicycle from $0.370$ to $0.482$ in the
ENGQ model. We also improved pedestrian segmentation in the generic SUN model.

It is interesting to note how these detection priors mix with geosemantic information. In the
presence of geometric context, pedestrian segmentation was slightly hurt. However, sitting pedestrian
segmentation is boosted from almost $0$ accuracy up to $0.108$. Bicycle also benefits from
geometric context and boosts from $0.458$ to $0.530$.

\paragraph{GIS-aware Segmentation}

To evaluate the influence of GIS features alone, we first trained a CRF model
without the image features from \cite{gould2009decomposing} and only used
the contextual features described in Section \ref{Segmentation} (GIS CRF). 
This yielded relatively good accuracy in the 4 labels present in the GIS model, but 
poor results for many others. This is quite natural since our GIS model does not include
detailed elements such as benches, and provides no information about what pixels might
be a bike or pedestrian on any given day. However, this model still improves a simple 
``blind'' backprojection of the GIS labels.

On the other hand, combining these GIS features with standard image features
gave a significant benefit, outperforming the image CRF baseline in almost all 
categories (+GIS rows in Figure \ref{fig:seg-table}). It is interesting to note
that labeling of some categories that did not appear in the GIS map data (e.g.
bench and wall) is still improved significantly by the geometric context provided
in the model (wall is boosted from $0.348$ to $0.481$, presumably since the
local appearance is similar to building but the geometric context is not).
This is in contrast to, e.g. \cite{Wang_2015_CVPR}, where all labels were included in
either the GIS or detector driven priors.


\paragraph{Scene Specific vs Generic Models}
Even without precisely resectioning the test image, there is a significant gain in
accuracy from knowing the rough camera location. When the camera location is
completely unknown, the best we can do is invoke the CRF trained on generic SUN
data ($0.309$ accuracy). However, if we know the camera is located somewhere on
the engineering quad, we can invoke the scene specific CRF trained on ENGQ to
boost performance to $0.561$. With resectioning, we can further utilize the 3D
model (+GIS) to gain an additional 3\% in segmentation performance by utilizing
geosemantic context features in the unary potential classifier. This gain is
quite significant given that some class baselines are already over $90\%$ in
accuracy leaving little room for improvement.


\begin{figure}[h]
\captionsetup[subfigure]{labelformat=empty}
\centering
\def\arraystretch{0.3}
\setlength{\tabcolsep}{0.3mm}
\hspace{-0.3cm}
\hspace{0.3cm}
\subfloat[GIS Backprojection]{%
\begin{tabular}{c}
	\includegraphics[width=0.3\columnwidth]{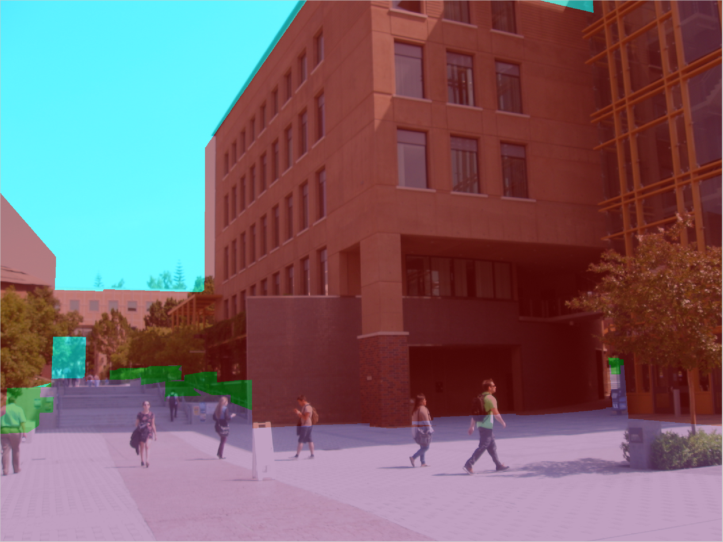} \\
	\includegraphics[width=0.3\columnwidth]{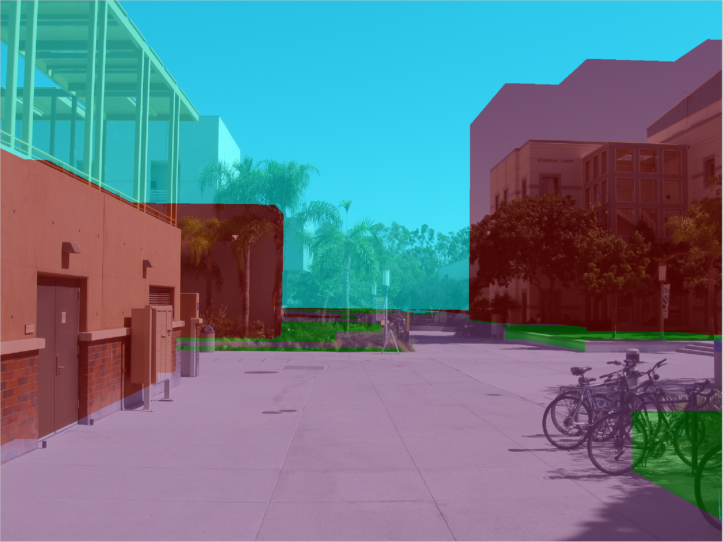} \\
	\includegraphics[width=0.3\columnwidth]{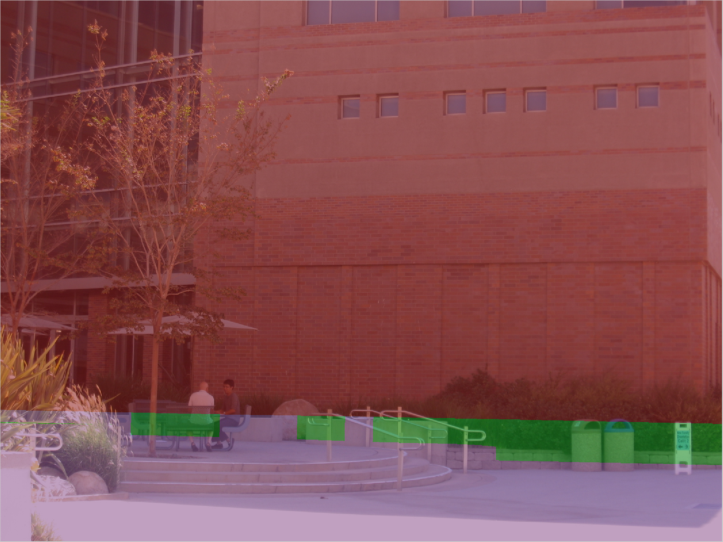} \\
	\includegraphics[width=0.3\columnwidth]{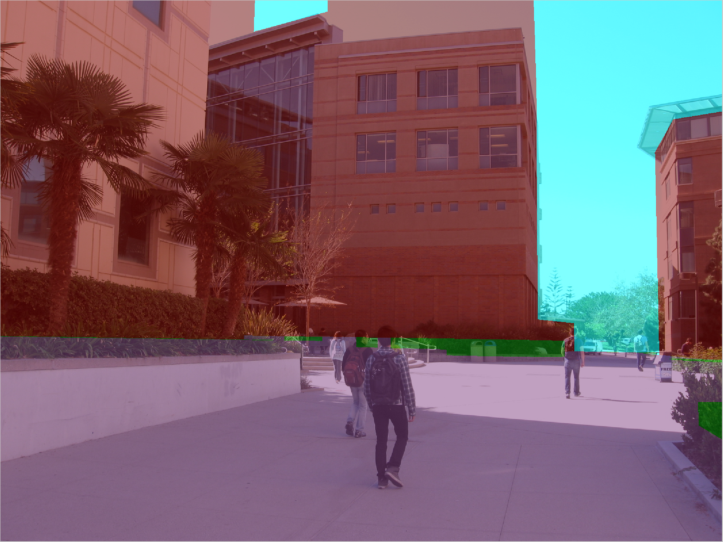} \\
\end{tabular} \hspace{-0.5cm}
}
\hspace{0.3cm}
\subfloat[Image CRF]{%
\begin{tabular}{c}
	\includegraphics[width=0.3\columnwidth]{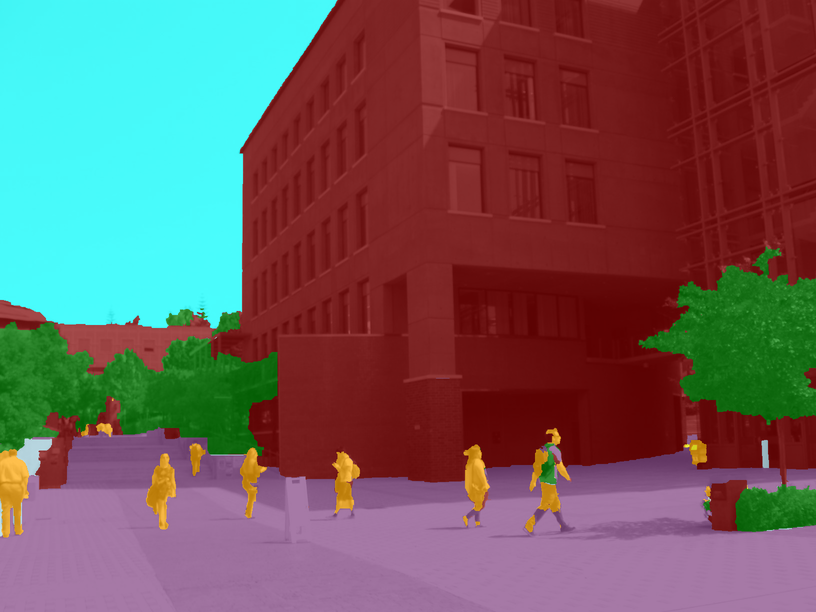} \\
	\includegraphics[width=0.3\columnwidth]{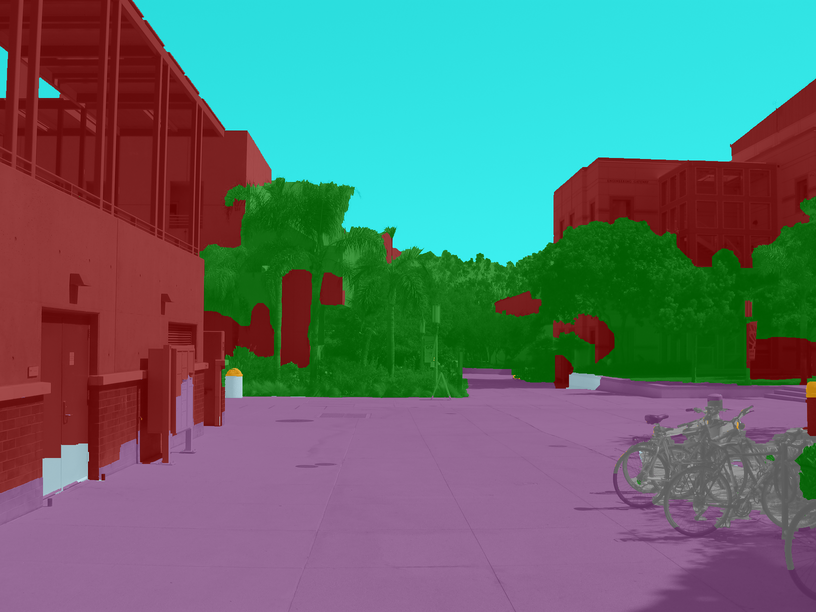} \\
	\includegraphics[width=0.3\columnwidth]{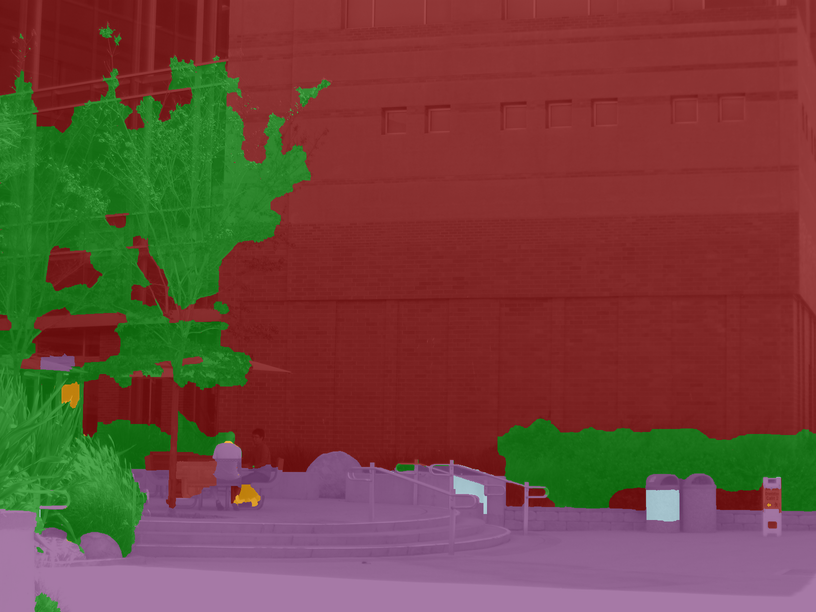} \\
	\includegraphics[width=0.3\columnwidth]{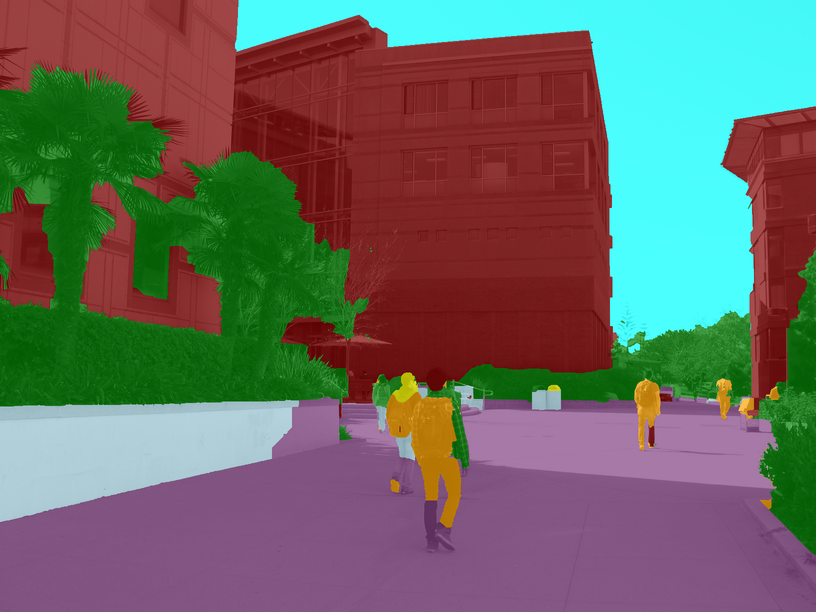} \\
\end{tabular} \hspace{-0.5cm}
}
\hspace{0.3cm}
\subfloat[Combined]{%
\begin{tabular}{c}
	\includegraphics[width=0.3\columnwidth]{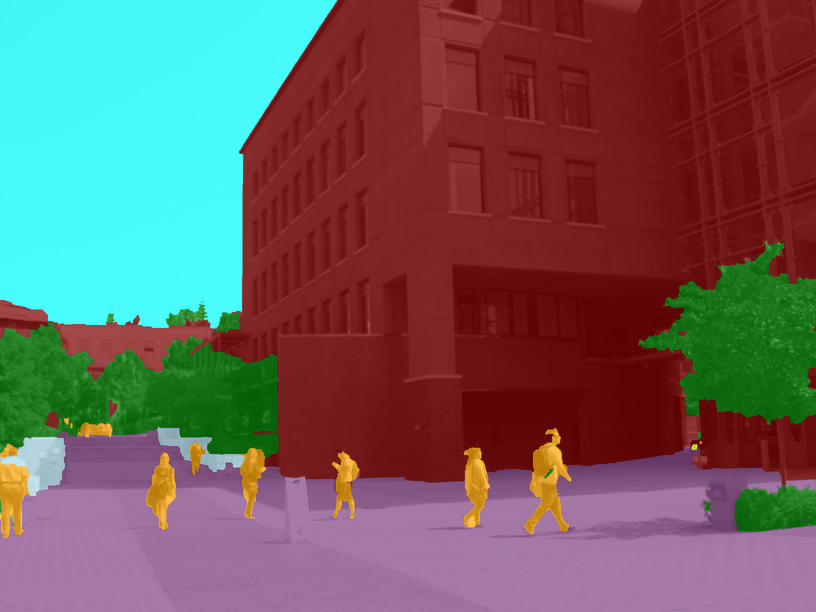} \\
	\includegraphics[width=0.3\columnwidth]{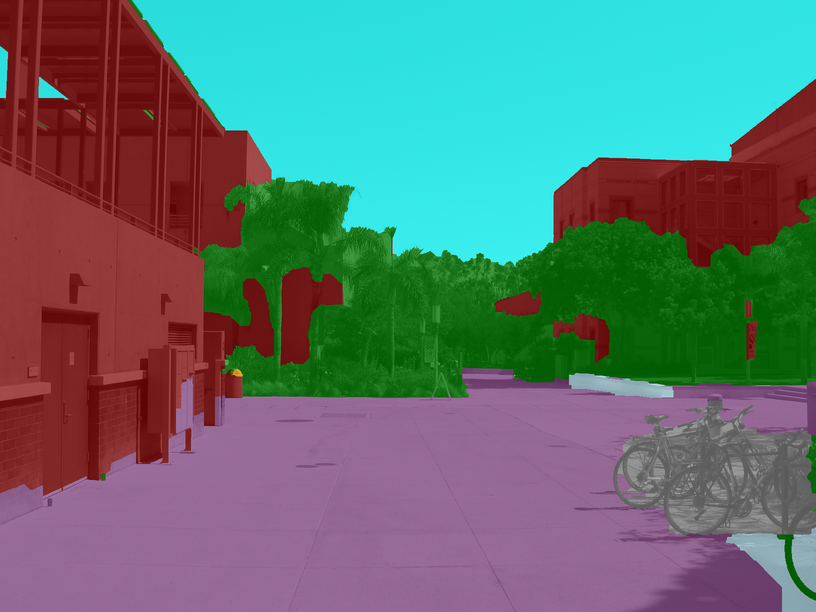} \\
	\includegraphics[width=0.3\columnwidth]{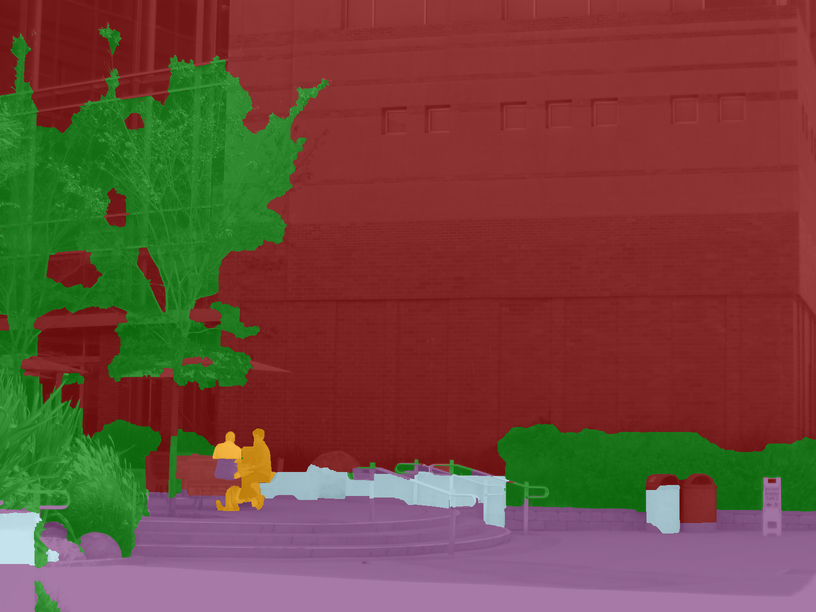} \\
	\includegraphics[width=0.3\columnwidth]{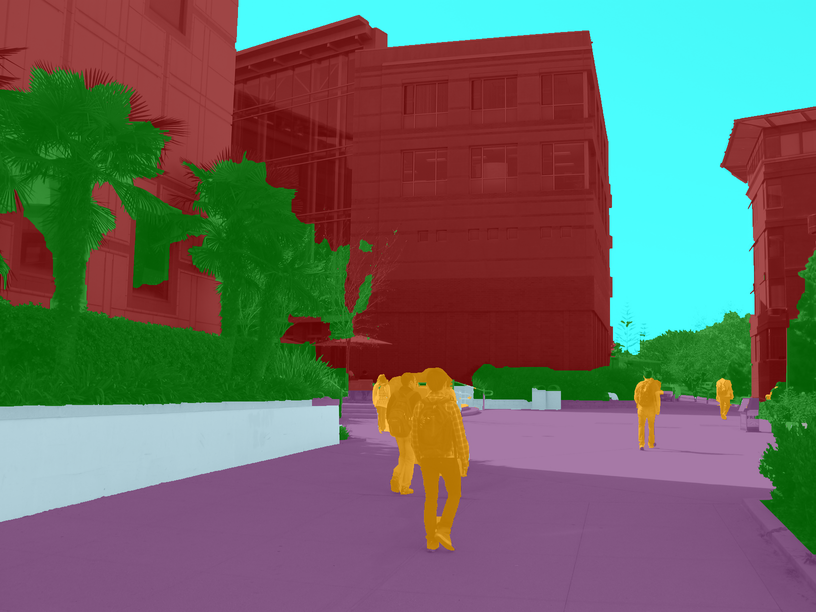} \\
\end{tabular} \hspace{-0.5cm}
}
\\
\vspace{-0.2cm}
\subfloat[]{
\begin{tabular}{l l l l l}
\textcolor[rgb]{0.502,0,0}{\rule{0.25cm}{0.25cm}} building &
\textcolor[rgb]{0,0.502,0}{\rule{0.25cm}{0.25cm}} plants &
\textcolor[rgb]{0.5020,0.2510,0.5020}{\rule{0.25cm}{0.25cm}} pavement &
\textcolor[rgb]{0,1,1}{\rule{0.25cm}{0.25cm}} sky &
\textcolor[rgb]{1,0.6471,0}{\rule{0.25cm}{0.25cm}} pedestrian \\
\textcolor[rgb]{1,1,0}{\rule{0.25cm}{0.25cm}} ped. sitting &
\textcolor[rgb]{0.5020,0.5020,0.5020}{\rule{0.25cm}{0.25cm}} bicycle &
\textcolor[rgb]{0.6471,0.1647,0.1647}{\rule{0.25cm}{0.25cm}} bench &
\textcolor[rgb]{0.6784,0.8471,0.902}{\rule{0.25cm}{0.25cm}} wall
\end{tabular} 
}

\caption{Qualitative segmentation results with overlaid images. Our combined model
improves over a image-based CRF by incorporating features derived from 
GIS (depth, labels, normals) and a DPM detector.}
\label{fig:seg_results}
\end{figure}

\section{Conclusion} \label{Discussion}

The rapid growth of digital mapping data in the form of GIS databases offers a
rich source of contextual information that should be exploited in practical
computer vision systems. We have described a basic pipeline that allows for
integration of such data to guide both traditional geometric reconstruction as
well as semantic segmentation and recognition. With a small amount of user
supervision, we can quickly lift 2D GIS maps into 3D models that immediately
provide strong scene geometry estimates (typically less than 5-10\% relative
depth error), greatly outperforming existing approaches monocular depth
estimation and providing a cheap alternative to laser range scanners. This
also provides strong geometric and semantic context features that can be 
exploited to improve detection and segmentation.

{\small
\bibliographystyle{ieee}
\bibliography{gis}
}

\end{document}